\documentclass[runningheads]{llncs}
\usepackage[T1]{fontenc}
\usepackage{graphicx}
\usepackage{subfigure}
\usepackage{xcolor}
\usepackage{amsmath}
\usepackage{amssymb}
\usepackage{hyperref}
\usepackage{orcidlink}
\begin{document}
\title{Patch-Level Glioblastoma Subregion Classification with a Contrastive Learning-Based Encoder\thanks{Corresponding author: Ying Weng. $^\dagger$ Juexin Zhang and Qifeng Zhong contributed euqally to this work.}}
\titlerunning{Patch-Level Glioblastoma Subregion Classification with Contrastive Encoder}
%
\author{Juexin Zhang$^\dagger$ \orcidlink{0000-0001-9086-7342} \and
Qifeng Zhong$^\dagger$ \orcidlink{0009-0004-9249-5027} \and
Ying Weng \orcidlink{0000-0003-4338-713X} \and
Ke Chen \orcidlink{0000-0002-2046-0034}
}
\authorrunning{J. Zhang et al.}
%
\institute{University of Nottingham Ningbo China, Ningbo 315100, China
\email{\{juexin.zhang,qifeng.zhong,ying.weng,ke.chen2\}@nottingham.edu.cn}\\}
\maketitle              
\begin{abstract}
The significant molecular and pathological heterogeneity of glioblastoma, an aggressive brain tumor, complicates diagnosis and patient stratification. While traditional histopathological assessment remains the standard, deep learning offers a promising path toward objective and automated analysis of whole slide images. For the BraTS-Path 2025 Challenge, we developed a method that fine-tunes a pre-trained Vision Transformer (ViT) encoder with a dedicated classification head on the official training dataset. Our model's performance on the online validation set, evaluated via the Synapse platform, yielded a Matthews Correlation Coefficient (MCC) of 0.7064 and an F1-score of 0.7676. On the final test set, the model achieved an MCC of 0.6509 and an F1-score of 0.5330, which secured our team second place in the BraTS-Pathology 2025 Challenge. Our results establish a solid baseline for ViT-based histopathological analysis, and future efforts will focus on bridging the performance gap observed on the unseen validation data.
\end{abstract}
\section{Introduction}

Glioblastoma is a highly aggressive primary brain tumor associated with poor patient outcomes, which makes accurate diagnosis and prognostic assessment critical for guiding therapy \cite{Zhang20129}. One of the main challenges lies in the marked histopathological heterogeneity of these tumors. Such heterogeneity is evident both across patients (inter-tumoral) and within individual tumors (intra-tumoral), and is reflected in differences in cellular morphology, phenotypic profiles, and treatment responses \cite{Piana2024}, thereby complicating prognostic evaluation.

Glioblastoma is an aggressive primary brain tumor associated with poor patient outcomes, which makes accurate diagnosis and prognostic prediction critical for guiding therapy \cite{Zhang20129}. The main challenge lies in the marked histopathological heterogeneity of these tumors. This diversity exists both between patients (inter-tumoral) and within a single tumor (intra-tumoral), manifesting as variations in cellular morphology, phenotypic expression, and therapeutic response \cite{Piana2024}, which makes the prognosis more complicated.

\begin{figure}[t]
    \centering
    \includegraphics[width=0.8\linewidth]{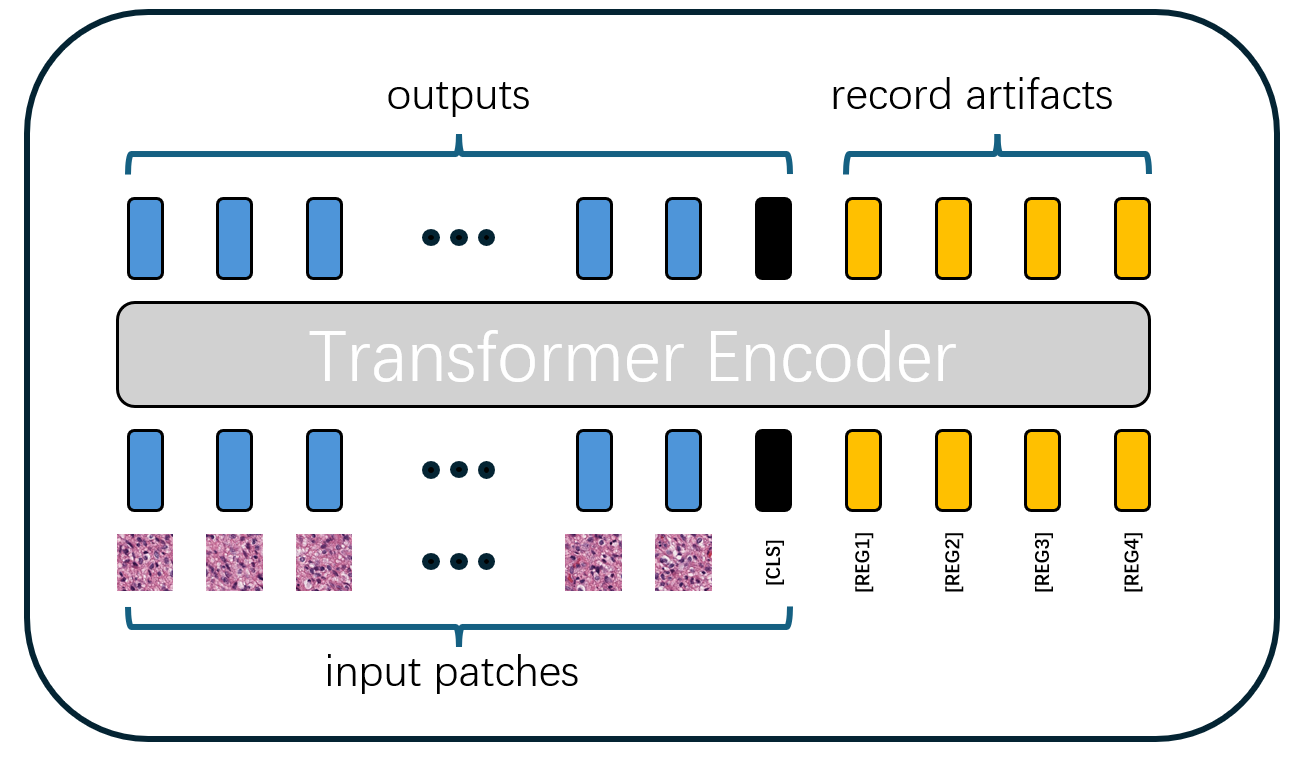}
    \caption{Virchow2 uses four registers to mitigate local information loss and enhance global contextual information by storing artifact tokens, thereby enabling better feature extraction. Adapted from \cite{darcet2024}.}
    \label{fig:1}
\end{figure}

The diagnosis of brain tumors has usually relied on pathology, where tissue samples are examined under a microscope to get subtypes of tumors and give treatment plans. In recent years, this traditional workflow has been reshaped by the rise of digital pathology, which replaces glass slides with high-resolution digital images of tissue sections. With the development of computational tools, digital pathology is redefining how brain tumors are studied and classified. The availability of whole-slide imaging also opens the door to artificial intelligence methods, especially deep learning models such as convolutional neural networks, which can detect subtle histological features and support faster, more accurate diagnoses.

Building upon this progress, large-scale foundation models pretrained on histopathology data, such as CTransPath \cite{WANG2022102559} and Virchow \cite{zimmermann202}. As reported by Neidlinger et al. \cite{neidlinger}, Virchow excels at extracting fine-grained features and outperforms many alternatives. We use the pretrained Virchow2 as a feature extraction backbone, leveraging its ability to capture discriminative tile-level features. For classification, we attach a head consisting of two linear layers to predict the class of each sub-region patch.

\section{Methods}

For our study, we selected Virchow2 \cite{zimmermann202} as the foundational pretrained model, chosen for its state-of-the-art performance in computational pathology. The architecture of Virchow2 is built upon the powerful Vision Transformer (ViT-H/14) \cite{DBLP:journals/corr/abs-2010-11929}, a high-capacity model known for its ability to capture complex spatial relationships in image data, as illustrated in Figure~\ref{fig:1}. The power of Virchow2 stems from its extensive pretraining on a massive and diverse dataset comprising approximately 3.1 million whole slide images (WSIs) from over 225,000 patients. This dataset provides comprehensive coverage of nearly 200 tissue types and multiple staining modalities, including the widely used hematoxylin and eosin (H\&E) as well as various immunohistochemical (IHC) stains. This robust pretraining ensures the model learns a rich hierarchy of generalizable features relevant to histopathology. Virchow2's feature representation capabilities were developed using DINOv2 \cite{oquab2024}, a sophisticated self-supervised learning framework. This approach enables the model to learn meaningful features directly from unlabeled images without requiring manual annotations.

\begin{figure}[t]
    \centering
    \includegraphics[width=1\linewidth]{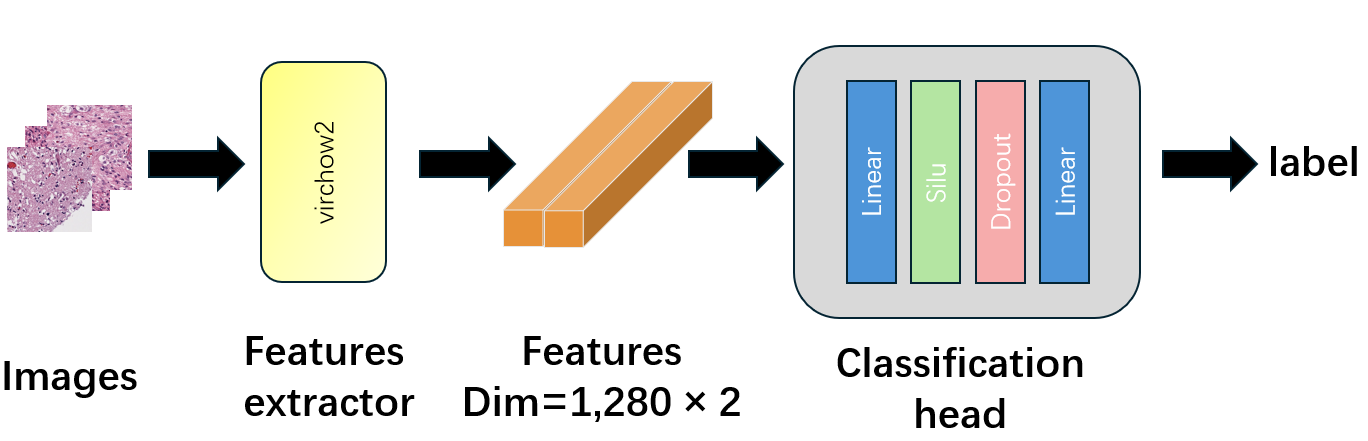}
    \caption{The network architecture of our model.}
    \label{fig:2}
\end{figure}

To tailor the pretrained Virchow2 for our specific 9-class sub-region classification task, we replace the original classification head with a new one, as depicted in Figure \ref{fig:2}. Our approach is designed to create a highly informative feature representation from each input patch before classification. The process begins by dividing each input patch into 256 non-overlapping tiles. These tiles are processed by the Virchow2 encoder, which generates 256 corresponding feature tokens, each with a dimensionality of 1280. To create a comprehensive feature representation for the entire patch, we combine two distinct signals:
\begin{itemize}
    \item Global Average Representation: We compute the mean of all 256 tokens to produce a single $1 \times 1280$ vector. This provides a global summary of the patch's overall texture, cellularity, and morphology.
    \item Semantic Class Token: We also retain the original class token ($1 \times 1280$), which the Vision Transformer learns to use as an aggregate representation of the most critical semantic information in the image.
\end{itemize}

The averaged feature token and the class token are concatenated and flattened to form a final, rich feature vector of size $2560$. This dual-representation strategy ensures our classifier has access to both a holistic summary and the most salient features identified by the transformer. This combined feature vector is then passed to our newly designed classification head, which consists of the following sequential layers:
\begin{itemize}
    \item A linear layer that projects the $2560$-dimensional input to a $256$-dimensional space, acting as a feature bottleneck to condense information.
    \item A SiLU (Sigmoid-weighted Linear Unit) activation function to introduce non-linearity, allowing the model to learn more complex decision boundaries.
    \item A dropout layer with a rate of $0.5$, which randomly deactivates neurons during training to effectively mitigate overfitting.
    \item A final linear output layer that maps the $256$-dimensional features to the 9 target class logits, producing the final predictions.
\end{itemize}

\section{Experiments}

\begin{figure}[t]
\centering
\subfigure[CT]{
\includegraphics[width=0.2\textwidth]{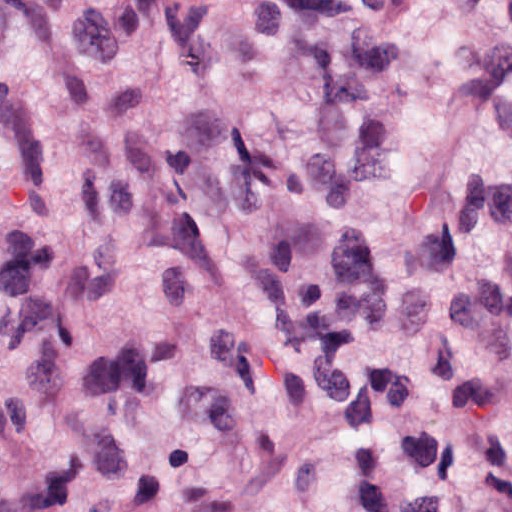}}
\hspace{0.05\textwidth}
\subfigure[DM]{
\includegraphics[width=0.2\textwidth]{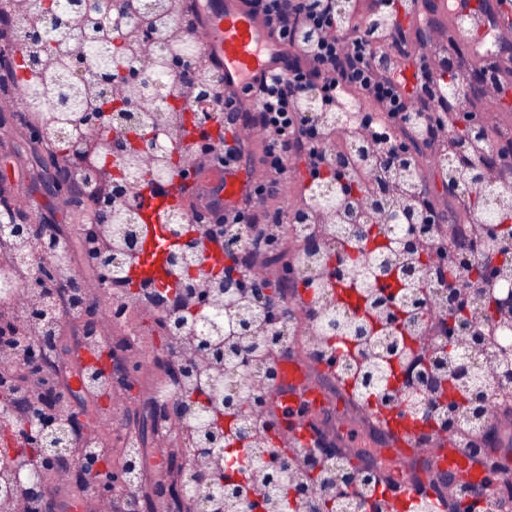}}
\hspace{0.05\textwidth}
\subfigure[IC]{
\includegraphics[width=0.2\textwidth]{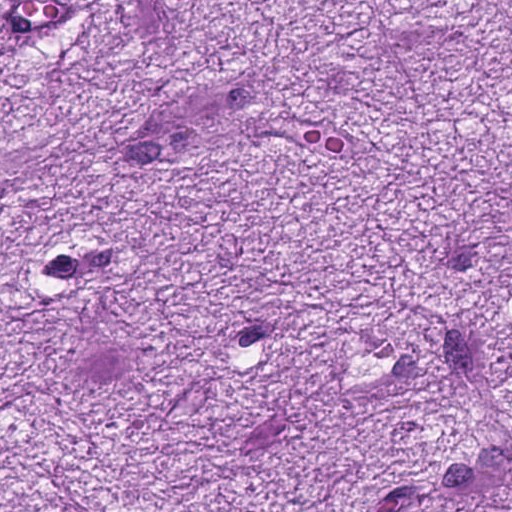}}

\subfigure[LI]{
\includegraphics[width=0.2\textwidth]{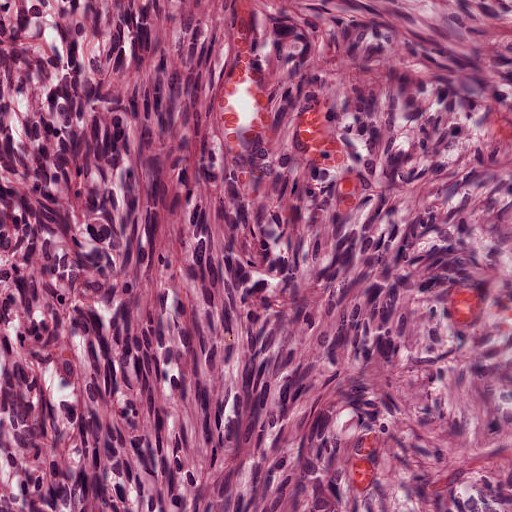}}
\hspace{0.05\textwidth}
\subfigure[MP]{
\includegraphics[width=0.2\textwidth]{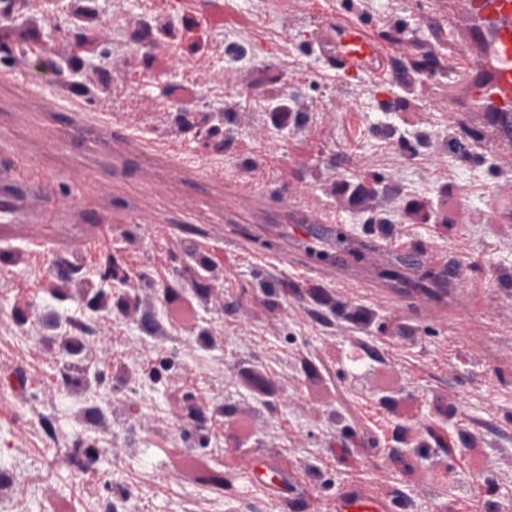}}
\hspace{0.05\textwidth}
\subfigure[NC]{
\includegraphics[width=0.2\textwidth]{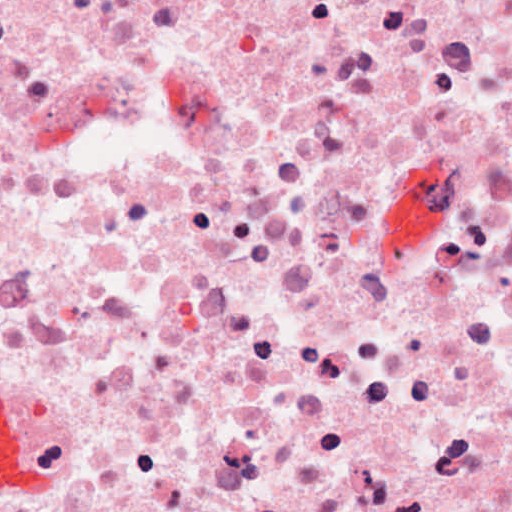}}

\subfigure[PL]{
\includegraphics[width=0.2\textwidth]{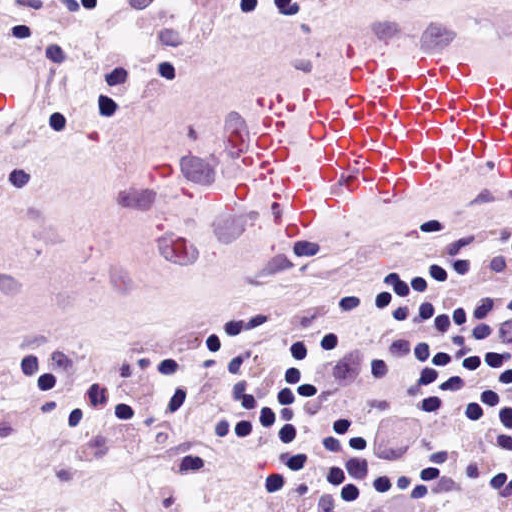}}
\hspace{0.05\textwidth}
\subfigure[PN]{
\includegraphics[width=0.2\textwidth]{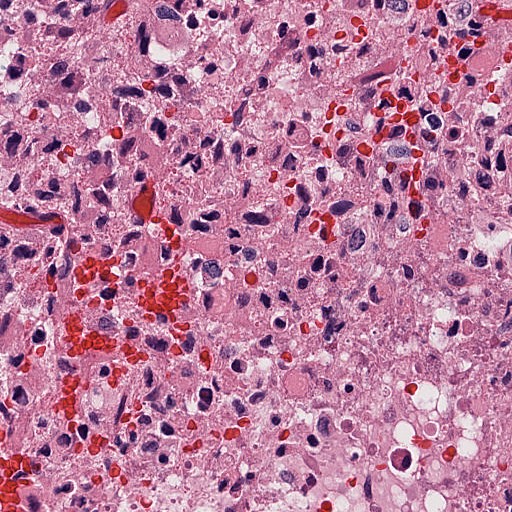}}
\hspace{0.05\textwidth}
\subfigure[WM]{
\includegraphics[width=0.2\textwidth]{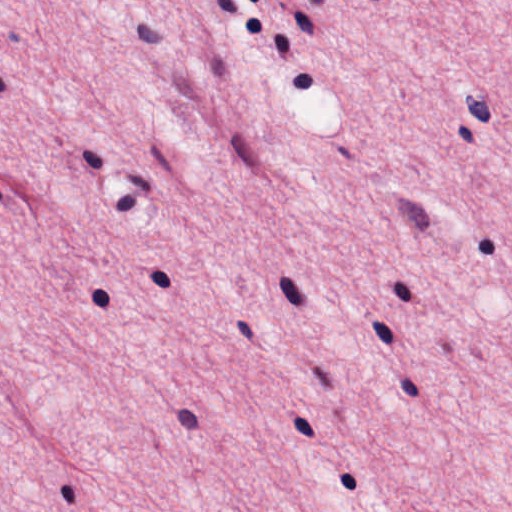}}

\caption{Illustration of the annotated histologic areas of interest.}
\label{fig:3}
\end{figure}

\subsection{Dataset}
\begin{figure}
    \centering
    \includegraphics[width=1\linewidth]{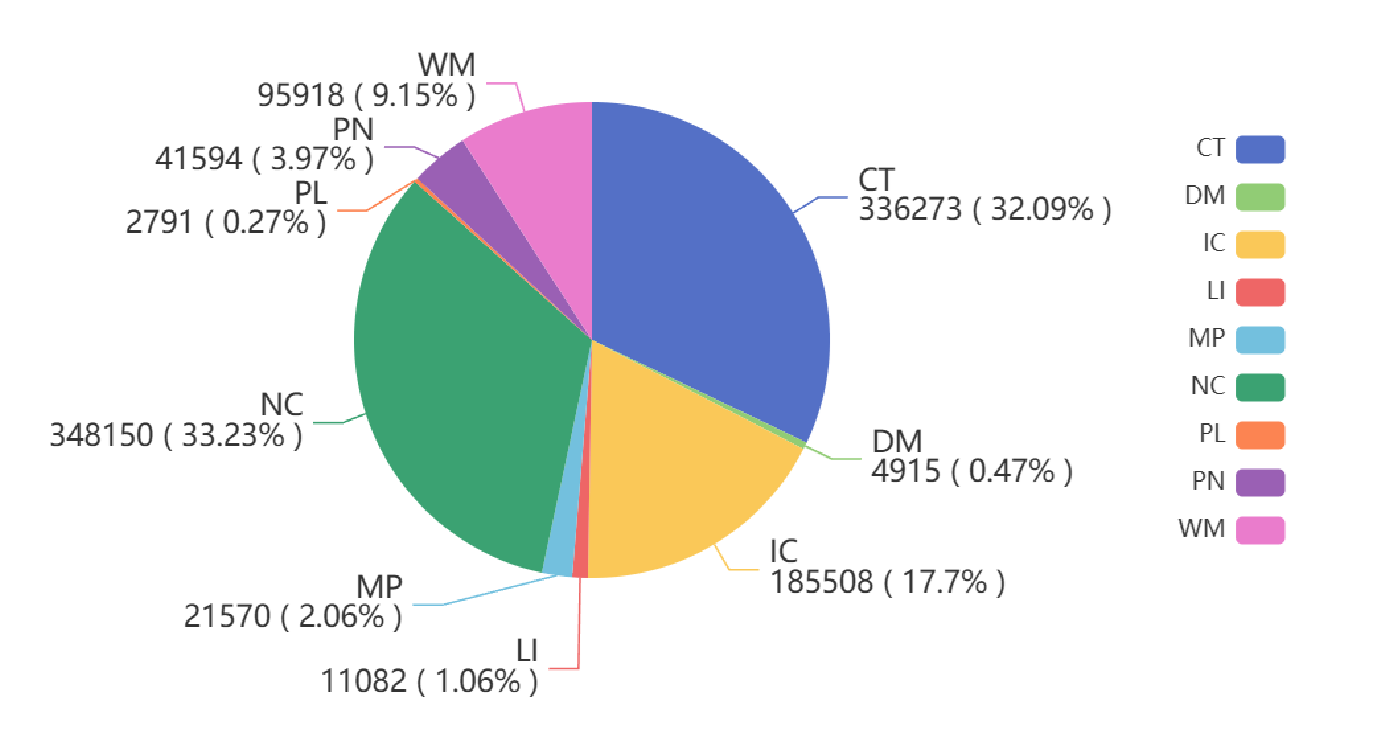}
    \caption{Class distribution of the dataset, detailing the number and percentage of samples for each category.}
    \label{fig:4}
\end{figure}

No external unlabeled pathology dataset was used, all operations in this paper were performed on the labeled BraTS-Path dataset \cite{bakas2024,Karargyris2023}, provided by the official BraTS 2025 challenge \cite{synapse2025}.
which consists of H\&E-stained Formalin-Fixed, Paraffin-Embedded (FFPE) tissue sections from the TCGA-GBM and TCGA-LGG collections. The data have been reclassified according to updated World Health Organization (WHO) criteria, focusing on glioblastoma cases. Expert annotations segment the slides into patches representing distinct histological regions. The dataset \cite{bakas2024,synapse2025} covers nine histological categories:
\begin{enumerate}
    \item Presence of cellular tumor (CT)
    \item Pseudopalisading necrosis (PN)
    \item Areas abundant in microvascular proliferation (MP)
    \item Geographic necrosis (NC)
    \item Infiltration into the cortex (IC)
    \item Penetration into white matter (WM)
    \item Leptomeningial infiltration (LI)
    \item Regions with dense macrophages (DM)
    \item Presence of lymphocytes (PL)
\end{enumerate}
Figure \ref{fig:3} provides a qualitative overview of the dataset with representative images from each of the nine classes. The dataset is highly unbalanced, exhibiting a long-tailed class distribution as quantitatively detailed in Figure \ref{fig:4}.

\subsection{Data pre-processing}
Histopathological patches were input data from nine classes. To ensure uniform input for training and evaluation and ensure the generalization ability of the model, all input images were transformed using a pre-processing pipeline implemented via:

\begin{itemize}
    \item \textbf{Resize}: All images were resized to a fixed size(resolution) of $224 \times 224$ pixels, ensuring uniform spatial dimensions of inputs.
    
    \item \textbf{ToTensor}: Images were converted into PyTorch tensors, scaling pixel values to the $[0, 1]$ range and rearranging the channel order to $(C, H, W)$.
    
    \item \textbf{Normalization}: Pixel intensities were normalized using ImageNet statistics with mean $[0.485,\ 0.456,\ 0.406]$ and standard deviation $[0.229,\ 0.224,\ 0.225]$, which facilitates convergence during training process.
\end{itemize}

\subsection{Evaluation Metrics}

Let $TP$, $TN$, $FP$, and $FN$ denote true positives, true negatives, false positives, and false negatives. The evaluation metrics used are defined as follows:

\begin{align}
&\text{Accuracy} = \frac{TP + TN}{TP + TN + FP + FN} \\[6pt]
&\text{Precision} = \frac{TP}{TP + FP} \\[6pt]
&\text{Recall} = \frac{TP}{TP + FN} \\[6pt]
&\text{F1\text{-}Score} = \frac{2 \cdot \text{Precision} \cdot \text{Recall}}{\text{Precision} + \text{Recall}} \\[6pt]
&\text{Specificity} = \frac{TN}{TN + FP} \\[6pt]
&\text{MCC} = \frac{TP \cdot TN - FP \cdot FN}{\sqrt{(TP + FP)(TP + FN)(TN + FP)(TN + FN)}}
\end{align}

A confusion matrix $C \in \mathbb{R}^{K \times K}$ was also used, where $C_{ij}$ represents the proportion of samples from class $i$ predicted as class $j$, and $K$ is the number of classes. All metrics were computed on the validation set.

\subsection{Experiment Settings}
A 5-fold stratified cross-validation \cite{10.5555/1643031.1643047} scheme was used to evaluate model generalizability. Let $\mathcal{D}_f^{\text{val}}$ denote the  validation sets for fold $f$, and let $\mathcal{M}_f$ be the model trained after fold $f$. The final performance metrics were obtained by averaging over the 5 validation folds:
\[
\text{Metric}_{\text{avg}} = \frac{1}{5} \sum_{f=1}^{5} \text{Metric}(\mathcal{M}_f, \mathcal{D}_f^{\text{val}})
\]

For each fold, we utilized the Adam optimizer with an initial learning rate of $1×e^{-5}$ and a weight decay of $0.01$. To ensure stable convergence, the learning rate schedule included a one-epoch warmup followed by a cosine annealing scheduler, reducing the learning rate to a minimum of $1×e^{-6}$. All models were trained on 4 NVIDIA V100 GPUs using a batch size of 256 and FP16 mixed-precision for computational efficiency.

\section{Results}

\subsection{Local Validation}
The model's performance was first evaluated on the local validation set using a 5-fold cross-validation methodology. A summary of key performance metrics for each class is provided in Table \ref{tab:metrics}. The results show a wide range of performance across classes. High-performing classes, such as NC, were identified with exceptional reliability, achieving scores above 95\% for all metrics. Moderately performing classes like IC and CT scored around 90\%, though CT exhibited lower precision (0.86) while IC had a comparatively low recall (0.82). For the remaining classes, while overall accuracy remained high, their recall and F1-scores were significantly lower, a trend particularly pronounced for LI, DM, and PL.

\begin{figure}[ht!]
    \centering
    \includegraphics[width=0.9\linewidth]{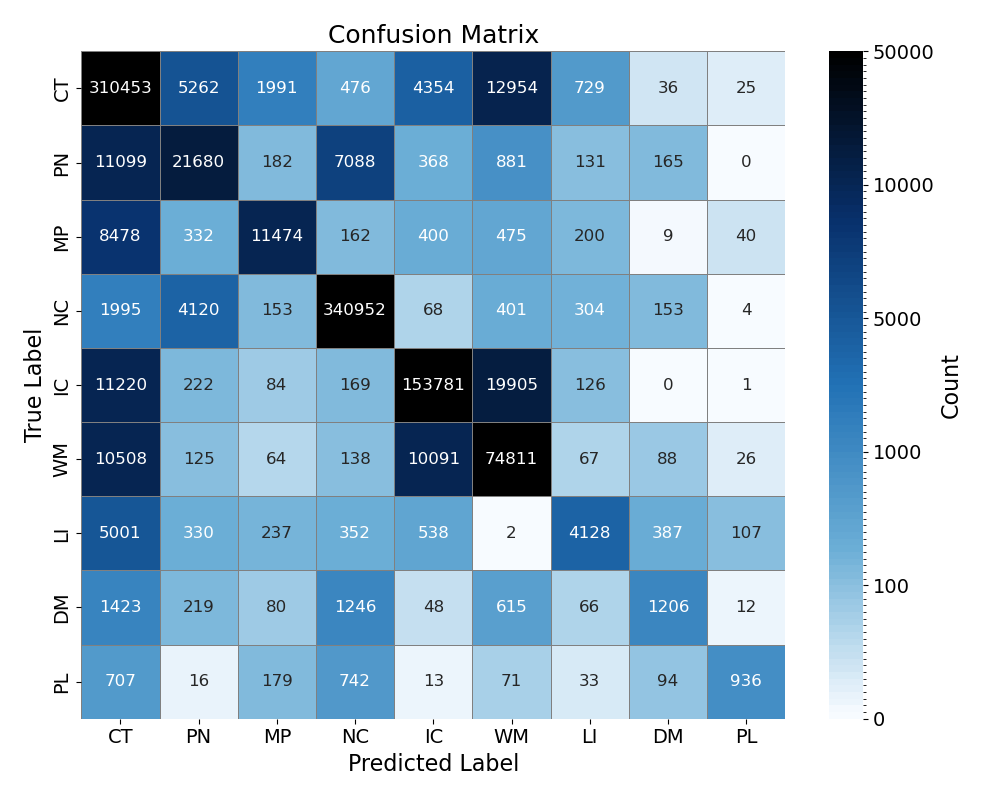}
    \caption{Aggregated confusion matrix from the 5-fold cross-validation on the local validation set.}
    \label{fig:5}
\end{figure}

\begin{table}[h!]
\centering
\caption{The class-wise performance on the local validation set. We calculate the micro average of different metrics.}
\label{tab:metrics}
\begin{tabular}{|c|c|c|c|c|c|c|c|c|c|c|c|}
\hline
Metric & CT & PN & MP & NC & IC & WM & LI & DM & PL & Average \\
\hline
Accuracy & 0.9272 & 0.9709 & 0.9875 & 0.9832 & 0.9546 & 0.9462 & 0.9918 & 0.9956 & 0.9980 & 0.8775 \\
\hline
Precision & 0.8603 & 0.6711 & 0.7944 & 0.9705 & 0.9064 & 0.6794 & 0.7137 & 0.5641 & 0.8132 & 0.8775 \\
\hline
Recall & 0.9232 & 0.5212 & 0.5319 & 0.9793 & 0.8290 & 0.7799 & 0.3725 & 0.2454 & 0.3354 & 0.8775 \\
\hline
Specificity & 0.9291 & 0.9894 & 0.9971 & 0.9852 & 0.9816 & 0.9629 & 0.9984 & 0.9991 & 0.9998 & 0.9847 \\
\hline
F1 & 0.8906 & 0.5867 & 0.6372 & 0.9749 & 0.8660 & 0.7262 & 0.4895 & 0.3420 & 0.4749 & 0.8775 \\
\hline
MCC & -- & -- & -- & -- & -- & -- & -- & -- & -- & 0.8347 \\
\hline
\end{tabular}
\end{table}

A deeper analysis of these results was conducted using the aggregated confusion matrix from all five folds, visualized in Figure \ref{fig:5}. The matrix reveals that CT, the majority class, is a primary source of confusion. The high concentration of false positives in its column indicates that samples from other classes are frequently misclassified as CT, which directly explains its lower precision. In contrast, NC, despite being the second-largest class, shows highly discriminative performance with minimal confusion.

The most notable trend is the model's difficulty with minority classes. The low recall values for LI, DM, and PL correspond directly to their small sample sizes in the dataset, confirming that the model struggles to learn their features and correctly identify them. Furthermore, the confusion matrix highlights other specific error patterns, such as the tendency for WM and IC to be misclassified as one another.

\subsection{Online Validation and Test Resluts}

The model's evaluation results, presented in Table~\ref{tab:online}, reveal a significant performance degradation when moving from the local to the online validation set. The model performed exceptionally well on the local data, achieving uniform scores of 0.8787 across Accuracy, Precision, Recall, and F1, with a corresponding MCC of 0.825. However, on the online validation set, these metrics declined sharply to 0.7677, with the F1-score reaching 0.5330 $\pm$ 0.036 and the MCC decreasing to 0.65087 $\pm$ 0.006. Such a marked discrepancy between the two validation environments is indicative of poor generalization and suggests a significant degree of overfitting.\cite{lin2018focallossdenseobject}.

\begin{table}[h!]
\centering
\caption{Online validation and test results.}
\label{tab:online}
\begin{tabular}{|l|c|c|c|c|c|c|}
\hline
  & Accuracy & Precision & Recall & Specificity & F1 & MCC \\
\hline
Validation set & 0.76766 & 0.76766 & 0.76766 & 0.97418 & 0.76766 &  0.70646 \\
\hline
Test set & N/A & N/A & N/A & N/A & 0.53301 & 0.65087\\
\hline
\end{tabular}
\end{table}

\section{Conclusion}

In this work, we presented a deep learning approach for the classification of glioblastoma histopathological subtypes by fine-tuning a pre-trained Virchow2 Vision Transformer. Our model demonstrated strong performance on a local validation set, achieving an F1-score of 0.8775 and an MCC of 0.8347, establishing it as a robust baseline for this complex classification task. The analysis revealed high accuracy for well-represented classes like Cellular Tumor (CT) and Necrosis (NC) but highlighted challenges with minority classes, which were often misclassified due to significant data imbalance.

In the BraTS-Pathology 2025 Challenge, our model achieved a second-place finish. It obtained an F1-score of 0.7676 and an MCC of 0.7064 on the Synapse online validation set, which decreased to 0.5330 and 0.6509, respectively, on the final test set. This performance discrepancy suggests that the model overfitted to the validation data, revealing a generalization gap. To address this, our future work will prioritize improving generalization by implementing advanced data augmentation and regularization strategies. Furthermore, we will investigate techniques specifically designed for long-tailed distributions to boost performance on less frequent histological classes.

\section*{Acknowledgements}
This work was supported by Ningbo Major Science \& Technology Project under Grant 2022Z126. (Corresponding: Ying Weng.)

\bibliographystyle{splncs04}
\bibliography{ref.bib}

\end{document}